\documentclass{article}
\usepackage{graphics}

\begin{document}

\title{A Novel Evolutionary Formulation of the Maximum Independent Set Problem}

\author{Valmir C. Barbosa\thanks{Corresponding author
({\tt valmir@cos.ufrj.br}).}\ \ and Luciana C. D. Campos\\
\\
Universidade Federal do Rio de Janeiro\\
Programa de Engenharia de Sistemas e Computa\c c\~ao, COPPE\\
Caixa Postal 68511\\
21941-972 Rio de Janeiro - RJ, Brazil}

\maketitle

\begin{abstract}
We introduce a novel evolutionary formulation of the problem of finding a
maximum independent set of a graph. The new formulation is based on the
relationship that exists between a graph's independence number and its acyclic
orientations. It views such orientations as individuals and evolves them with
the aid of evolutionary operators that are very heavily based on the structure
of the graph and its acyclic orientations. The resulting heuristic has been
tested on some of the Second DIMACS Implementation Challenge benchmark graphs,
and has been found to be competitive when compared to several of the other
heuristics that have also been tested on those graphs.

\end{abstract}

\section{Introduction}\label{intr}

Let $G=(N,E)$ be an undirected graph of node set $N$ and edge set $E$ such that
$n=\vert N\vert$ and $m=\vert E\vert$. An {\it independent set\/} (or
{\it stable set\/}) of $G$ is a subset $S$ of $N$ containing no
{\it neighbors\/} (nodes that are connected by an edge in $G$). The set
$N\setminus S$ is a {\it node cover\/} of $G$, that is, a set of nodes that
includes at least one of the two end nodes of every edge. We call every subset
of $N$ whose nodes are all neighbors of one another in $G$ a {\it clique}. The
{\it complement\/} $\bar G$ of $G$ is an undirected graph of node set $N$ in
which two nodes are neighbors if and only if they are not neighbors in $G$.
Clearly, $S$ is an independent set of $G$ if and only if it is a clique of
$\bar G$.

We are concerned in this paper with the problem of finding a maximum independent
set in $G$, that is, an independent set of maximum size. Equivalently, this
problem can be viewed as asking for a minimum node cover in $G$ or a maximum
clique in $\bar G$. Finding independent sets (or any of the other equivalent
structures) of extremal size has several important applications. We refer the
reader to the volume that resulted from the Second DIMACS Implementation
Challenge \cite{jt96} for various examples of application areas, and to
\cite{eo98} for further examples from coding theory.

The problem of finding a maximum independent set in $G$ is NP-hard (it is
NP-complete when formulated as a decision problem \cite{k72,gj79}), and remains
NP-hard even if we settle for solving it approximately within $n^{1/4-\epsilon}$
of the optimum for any $\epsilon>0$ \cite{al97,acgkmp99,bgs98}. That is, if
$\alpha(G)$ is the size of a maximum independent set of $G$ (the
{\it independence number\/} of $G$), then finding an independent set of size at
least $n^{1/4-\epsilon}\alpha(G)$ is NP-hard.

Our formulation of the maximum independent set problem is based on the notion
of an {\it acyclic orientation\/} of $G$, i.e., an assignment of directions to
the edges of $G$ that leads to no directed cycles. Let $\Omega(G)$ denote the
set of all the acyclic orientations of $G$. For $\omega\in\Omega(G)$, let
$D_\omega$ be the set of all {\it chain decompositions\/} of the nodes of $G$
according to $\omega$, that is, each member of $D_\omega$ is a partition of $N$
into sets that correspond to chains (directed paths) according to $\omega$. For
$d\in D_\omega$, let $\vert d\vert$ denote the number of chains in $d$. Our
point of departure is the following equality, which relates the independence
number of $G$ to its acyclic orientations \cite{d79}:
\begin{equation}
\label{alphafromomega}
\alpha(G)=\max_{\omega\in\Omega(G)}\min_{d\in D_\omega}\vert d\vert.
\end{equation}

\begin{figure}[t]
\centering
\scalebox{1.0}{\includegraphics{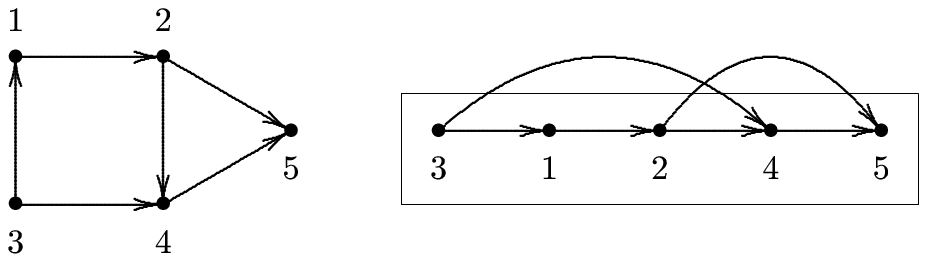}}\\
\vspace{0.1in}
\scalebox{1.0}{\includegraphics{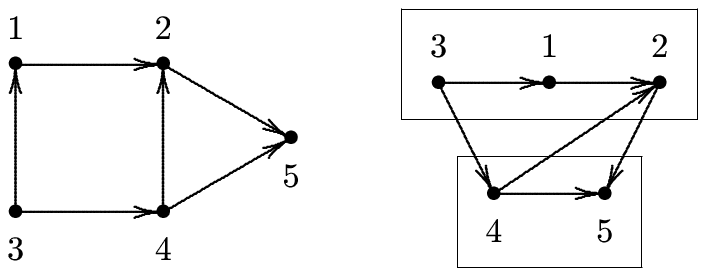}}
\caption{Two acyclic orientations and the corresponding minimum chain
decompositions.}
\vspace{0.2in}
\label{width}
\end{figure}

By (\ref{alphafromomega}), $\alpha(G)$ is the number of chains in the chain
decomposition of $N$ into the fewest possible chains, according to the acyclic
orientation of $G$ for which that number is greatest. This result is a
refinement of Dilworth's theorem \cite{d50} and is illustrated in
Figure~\ref{width}, where two acyclic orientations of the same graph are shown
alongside the corresponding minimum chain decompositions. In the figure, the
bottommost acyclic orientation is the one whose minimum chain decomposition is
greatest, thence $\alpha(G)=2$ for the graph in question.

As we see it, the greatest significance of (\ref{alphafromomega}) is that it
spells out how the set $\Omega(G)$ can be regarded as a set of individuals, the
fittest of which yields the independence number of $G$.\footnote{A relation dual
to the one in (\ref{alphafromomega}) indicates how to express the chromatic
number of $G$ (cf.\ \cite{b98}) in terms of its acyclic orientations \cite{d79}.
The use of that other relation in an evolutionary approach to find the graph's
chromatic number has been developed by one of us and collaborators
\cite{ban03}.} Viewing the maximum independent set problem from this perspective
is based on taking $\min_{d\in D_\omega}\vert d\vert$ as the measure of fitness
for individual $\omega$ and on searching $\Omega(G)$ for an individual of
maximum fitness.\footnote{$\Omega(G)$ is a set of vast dimensions. For example,
the number of distinct acyclic orientations of $G$ is $2^m$ if $G$ is a tree,
$n!$ if it is a complete graph (all nodes connected to all others), and $2^n-2$
if it is a ring. In general, and remarkably, the number of members of
$\Omega(G)$ is given by the absolute value of the chromatic polynomial of $G$
(cf.\ \cite{b98}) applied to the negative unit \cite{s73}.}

We develop this evolutionary approach in the remainder of the paper, starting
in Section~\ref{formul}, where the details of the new formulation are
introduced, including how to compute an individual's fitness and the
evolutionary operators of crossover and mutation for $\Omega(G)$. We then
continue in Section~\ref{testset} with a description of the graphs to be used
in our experiments, whose results are reported in Section~\ref{results}.
Conclusions are given in Section~\ref{concl}.

\section{The formulation}\label{formul}

\begin{figure}[t] 
\centering
\scalebox{1.0}{\includegraphics{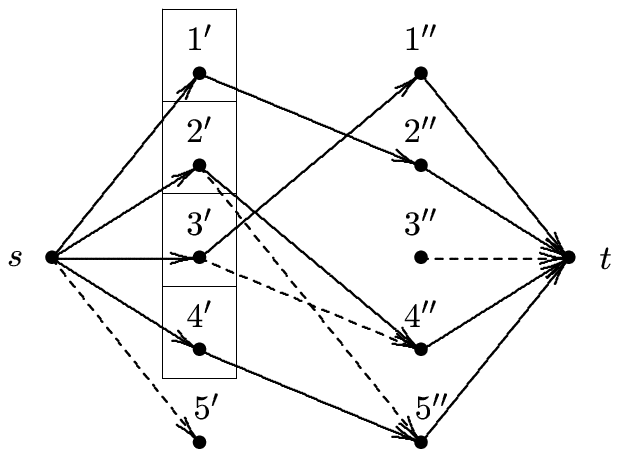}}\\
\vspace{0.1in}
\scalebox{1.0}{\includegraphics{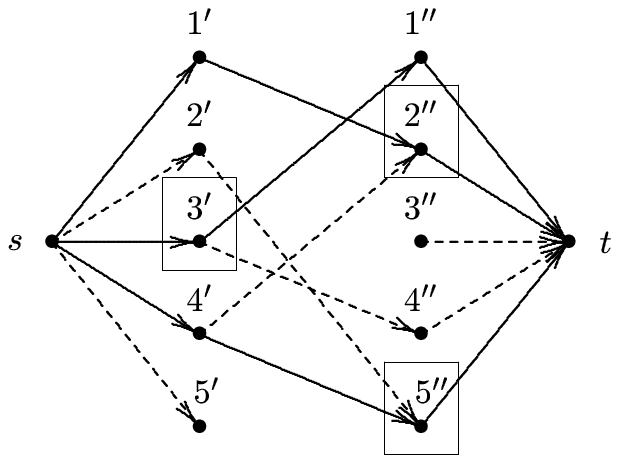}}
\caption{Flow networks associated with the acyclic orientations of
Figure~\ref{width}.}
\vspace{0.2in}
\label{flow}
\end{figure}

The key to a better understanding of how to use (\ref{alphafromomega}) is that
$\min_{d\in D_\omega}\vert d\vert$, the purported measure of fitness for
individual $\omega\in\Omega(G)$, is in fact the size of an independent set that
can be unequivocally obtained from orientation $\omega$. In order to see this,
we associate a directed graph, call it $D(G,\omega)$, with $G$ oriented by
$\omega$. This directed graph has $2n+2$ nodes: two distinguished nodes, called
$s$ and $t$, and two nodes, called $i'$ and $i''$, for each node $i\in N$.
For every $i\in N$, $D(G,\omega)$ has an edge directed from $s$ to $i'$ and
another from $i''$ to $t$. For every $(i,j)\in E$ that is directed by $\omega$
from $i$ to $j$, in $D(G,\omega)$ an edge exists from $i'$ to $j''$. Node $s$ is
therefore a {\it source\/} (a node with no edges directed inward) and node $t$ a
{\it sink\/} (a node with no edges directed outward). An illustration of this
construction is given in Figure~\ref{flow}, where the directed graphs
corresponding to the two acyclic orientations of Figure~\ref{width} are shown.

We now regard $D(G,\omega)$ as a flow network whose edges either have unit
capacity (those leaving $s$ or arriving at $t$) or infinite capacity (all
others), and consider the maximum flow from $s$ to $t$, whose value we assume
to be $F$. This flow is necessarily integral, as illustrated in
Figure~\ref{flow}, where solid lines have been used to draw edges carrying unit
flow and dashed lines those carrying zero flow. It also establishes a chain
decomposition of $G$ according to $\omega$ into $n-F$ chains. Take, for example,
the topmost network of Figure~\ref{flow}, where $F=4$. If we only follow edges
carrying unit flow, then we can easily trace the single chain $3'\to 1''$,
$1'\to 2''$, $2'\to 4''$, $4'\to 5''$. For the bottommost network, $F=3$ and we
get the two chains $3'\to 1''$, $1'\to 2''$ and $4'\to 5''$. These, readily, are
the minimum chain decompositions according to the two acyclic orientations.

What is left to note is that only edges leaving $s$ or arriving at $t$ can be
saturated by the maximum flow, so the corresponding minimum cut is necessarily
given by a group of such edges. In Figure~\ref{flow}, the minimum cut in each
network is indicated by boxes enclosing the nodes other than $s$ or $t$ that are
involved in the cut. Such nodes do necessarily constitute a minimum node cover
of the edges that do not involve $s$ or $t$ in $D(G,\omega)$, and consequently
induce a node cover in $G$ as well ($\{1,2,3,4\}$ for the topmost network,
$\{2,3,5\}$ for the other) with a corresponding independent set ($\{5\}$ and
$\{1,4\}$, respectively).

In general, then, we have the following \cite{amo93}. If $F$ is the value of the
maximum flow from $s$ to $t$ in $D(G,\omega)$, then $n-F$ is the number of
chains in the minimum chain decomposition of $G$ according to $\omega$ and $F$
is the size of a node cover in $G$. Consequently, $n-F$ is also the size of an
independent set in $G$. In order to determine the actual nodes that constitute
this independent set, it suffices to look at the minimum cut in $D(G,\omega)$
and at the node cover it induces on the edges of $D(G,\omega)$ that do not touch
$s$ or $t$. This node cover corresponds to a node cover in $G$ as well, whose
complement with respect to $N$ is then the desired independent set.

We now turn to the three key elements of our formulation, namely how to assess
an individual's fitness and how crossover and mutation operate. We have designed
these elements in such a way that they can be used directly in most standard
templates of fitness-maximization genetic algorithms \cite{g89,m96}. What is
novel in our formulation is the adoption of $\Omega(G)$ as the search space out
of which populations are formed. Not only do the acyclic orientations of $G$
relate cleanly, as we have discussed, to the independent sets of $G$, but also
they allow for evolutionary operators that are simple and yet effective in
several aspects of the evolutionary search. We will come to them shortly.

\subsection*{Fitness evaluation}

For $\omega\in\Omega(G)$, let $f(\omega)$ denote the fitness to be maximized
over $\Omega(G)$ while searching for a maximum independent set of $G$. By
(\ref{alphafromomega}), we have
\begin{equation}
\label{fitnessfromomega}
f(\omega)=\min_{d\in D_\omega}\vert d\vert,
\end{equation}
that is, the fitness of individual $\omega$ is the number of chains in the
minimum chain decomposition according to that orientation.

It follows from our preceding discussion that $f(\omega)$ can be assessed along
the following steps:
\begin{enumerate}
\item Construct the flow network $D(G,\omega)$.
\item Compute the value $F$ of the maximum flow from $s$ to $t$ in
$D(G,\omega)$.
\item Let $f(\omega)=n-F$.
\end{enumerate}

We note, with regard to Step~2, that only the value $F$ of the maximum flow is
needed for fitness evaluation, not the actual flow. In other words, what is
needed is the maximum total flow incoming to $t$, not the particular assignment
of flows to all edges. When a push-relabel method is used to compute the maximum
flow, it is a simple matter to separate the computation into two phases
\cite{gt88}: the first phase computes $F$ but may leave excess flow at some
nodes; the second phase corrects this by returning flow in order to eliminate
excesses. Conveniently, one of the most successful implementations currently
available of a maximum-flow algorithm does precisely this \cite{cg97}, and as
such allows the computation to stop at the end of the first phase, right after
$F$ has been found. What is also convenient is that, when push-relabel methods
are thus implemented, at the end of the first phase the minimum cut is also
known, which is useful for determining the independent set that corresponds to
the best individual found during the evolutionary search, in the manner we
indicated earlier in this section. We return to this in Section~\ref{results}.

\subsection*{Crossover}

The crossover of the two individuals $\omega_1,\omega_2\in\Omega(G)$ to yield
the two offspring $\omega'_1,\omega'_2\in\Omega(G)$ is best described in terms
of a linear representations of the individuals involved. For individual
$\omega\in\Omega(G)$, the representation we adopt, denoted by $L(\omega)$, is
the sequence $L(\omega)=\langle i_1,\ldots,i_n\rangle$, where $i_1,\ldots,i_n$
are the nodes of $G$. In this sequence, and for $1\le x,y\le n$, node $i_x$
appears to the left of $i_y$ (i.e., $x<y$) if $(i_x,i_y)$ is an edge of $G$ and
is oriented by $\omega$ from $i_x$ to $i_y$. For example, both
$\langle 3,1,4,2,5\rangle$ and $\langle 3,4,1,2,5\rangle$ are valid linear
representations of the bottommost acyclic orientation of Figure~\ref{width}. 
Clearly, $L(\omega)$ represents $\omega$ unambiguously, though not uniquely.

Now let $L(\omega_1)=\langle i_1,\ldots,i_n\rangle$ and
$L(\omega_2)=\langle j_1,\ldots,j_n\rangle$. Let also $z$ such that $1\le z<n$
be the crossover point. Then $L(\omega'_1)=\langle i'_1,\ldots,i'_n\rangle$ and
$L(\omega'_2)=\langle j'_1,\ldots,j'_n\rangle$, where
\begin{itemize}
\item $\langle i'_1,\ldots,i'_z\rangle=\langle i_1,\ldots,i_z\rangle$;
\item $\langle i'_{z+1},\ldots,i'_n\rangle$ is the subsequence of
$\langle j_1,\ldots,j_n\rangle$ comprising all nodes that are not in
$\langle i_1,\ldots,i_z\rangle$;
\item $\langle j'_1,\ldots,j'_z\rangle=\langle j_1,\ldots,j_z\rangle$;
\item $\langle j'_{z+1},\ldots,j'_n\rangle$ is the subsequence of
$\langle i_1,\ldots,i_n\rangle$ comprising all nodes that are not in
$\langle j_1,\ldots,j_z\rangle$.
\end{itemize}

Notice that the $L(\omega'_1)$ and $L(\omega'_2)$ thus determined are valid
linear representations of acyclic orientations, since, by construction, both
sequences contain all nodes from $G$. Furthermore, $\omega'_1$ and $\omega'_2$
inherit edge orientations from $\omega_1$ and $\omega_2$ as follows:
\begin{itemize}
\item Edges joining nodes in the set $\{i'_1,\ldots,i'_z\}$ to any other nodes
are oriented by $\omega'_1$ exactly as by $\omega_1$.
\item Edges joining nodes in the set $\{i'_{z+1},\ldots,i'_n\}$ exclusively are
oriented by $\omega'_1$ exactly as by $\omega_2$.
\item Edges joining nodes in the set $\{j'_1,\ldots,j'_z\}$ to any other nodes
are oriented by $\omega'_2$ exactly as by $\omega_2$.
\item Edges joining nodes in the set $\{j'_{z+1},\ldots,j'_n\}$ exclusively are
oriented by $\omega'_2$ exactly as by $\omega_1$.
\end{itemize}

\subsection*{Mutation}

Like crossover, our mutation operator is defined in terms of the linear
representations of individuals. We use single-locus mutation at the mutation
point $z$ with $1\le z\le n$. For individual $\omega\in\Omega(G)$, the mutation
operator turns node $i_z$ into a source, thus yielding another acyclic
orientation $\omega'$. That $\omega'$ is indeed acyclic has been argued
elsewhere \cite{b00}: in essence, a directed cycle through $i_z$ would be
required for $\omega'$ not to be acyclic, which is impossible, $i_z$ being a
source according to $\omega'$.

We also refer the reader elsewhere (\cite{ban03}, Section~4) for an
argumentation as to why this type of mutation does more for the evolutionary
search than simply to allow occasional random jumps loosely intended to escape
local optima. Specifically, what is shown is that, for any two acyclic
orientations $\omega,\omega'\in\Omega(G)$, there necessarily exists a finite
sequence of mutations that turns $\omega$ into $\omega'$. So this operator can
be regarded as providing the search space $\Omega(G)$ with an underlying
fundamental connectedness that allows, at least in principle, every acyclic
orientation to be reached regardless of where the evolutionary search is
started.

\section{The experimental test set}\label{testset}

In this section we give a brief description of the benchmark graphs used in the
experiments reported in Section~\ref{results}. They have all been extracted from
the DIMACS challenge suite \cite{t96}. That suite is structured from the
perspective of finding maximum cliques, so the graph types listed next
characterize $\bar G$.

\bigskip
\noindent
{\tt c-fat$n$-$c$} \cite{hpv93}. This graph comes from fault-diagnosis problems
\cite{bp90} and its set of $n$ nodes is partitioned into $n/c\log n$ sets of
approximately equal sizes. Edges are deployed so that every node is connected
to every other node in its own set and in the two sets that are neighbors to its
own (according to an arrangement of the sets into a ring).

\bigskip
\noindent
{\tt johnson$W$-$w$-$d$} \cite{hpv93}. This graph arises in problems from
coding theory. It has $n={W\choose w}$ nodes, each node labeled with a $W$-digit
binary number having exactly $w$ $1$'s. Two nodes are joined by an edge if the
Hamming distance between their labels (the number of digits at which they
differ) is at least $d$.

\bigskip
\noindent
{\tt keller$d$} \cite{ls92}. This graph is derived from another with $4^d$ nodes
which arises in connection with proving Keller's 1930 conjecture false for high
dimensions. The conjecture is that a ``tiling'' of Euclidean $d$-dimensional
space by unit cubes necessarily contains two cubes meeting in a full
$(d-1)$-dimensional face.

\bigskip
\noindent
{\tt hamming$W$-$d$} \cite{hpv93}. This graph comes from coding-theory problems
also and has $n=2^W$ nodes, each node labeled with a $W$-digit binary number.
Two nodes are connected if the Hamming distance between their labels is at least
$d$.

\bigskip
\noindent
{\tt san$n$\_$f$\_x} \cite{jsg96}. This graph is artificially constructed on $n$
nodes to have a maximum clique whose size is determined beforehand. Its number
of edges is $fn(n-1)/2$; {\tt x} is only used to differentiate among instances.

\bigskip
\noindent
{\tt sanr$n$\_$p$} \cite{jsg96}. This is a random graph on $n$ nodes,
generated by adding an edge with constant probability $p$ between any two
distinct nodes \cite{b01}. It is expected to have dimensions close to those of
{\tt san$n$\_$f$\_x} for $f=p$.

\bigskip
\noindent
{\tt brock$n$\_x} \cite{bc96}. This is a random graph on $n$ nodes, designed
to have a maximum clique much larger than would be expected from the nodes'
degrees. The number {\tt x} is used for instance differentiation.

\bigskip
\noindent
{\tt p\_hat$n$-x} \cite{sg93}. This is a random graph on $n$ nodes whose density
is based on two parameters. This contrasts with the usual random graphs with
fixed edge probability \cite{b01}, leading to node degrees that are more
spread and to larger cliques also. The number {\tt x} differentiates among
instances.

\bigskip
\noindent
{\tt MANN\_ax} \cite{ms96}. This graph gives the clique formulation of the
instance $A{\tt x}$ of the Steiner triple problem. This formulation is obtained
by a conversion from the set-covering formulation of that problem \cite{ms95}.

\bigskip
Details on the graphs we used in our experiments are given in Tables
\ref{graphs1} and \ref{graphs2}. For each $\bar G$, the tables give the values
of $n$ and $m$ (the number of edges in $G$), as well as $\alpha(G)$, when known
from design characteristics.

\begin{table}[t]
\centering
\caption{Benchmark graphs.}
\begin{tabular}{lrrr}
$\bar G$			&$n$	&$m$		&$\alpha(G)$	\\
\hline
{\tt c-fat$200$-$1$}		&$200$	&$18{,}366$	&		\\
{\tt c-fat$200$-$2$}		&$200$	&$16{,}665$	&		\\
{\tt c-fat$200$-$5$}		&$200$	&$11{,}427$	&		\\
{\tt c-fat$500$-$1$}		&$500$	&$120{,}291$	&		\\
{\tt c-fat$500$-$2$}		&$500$	&$115{,}611$	&		\\
{\tt c-fat$500$-$5$}		&$500$	&$101{,}559$	&		\\
{\tt c-fat$500$-$10$}		&$500$	&$78{,}123$	&		\\
{\tt johnson$8$-$2$-$4$}	&$28$	&$168$		&		\\
{\tt johnson$8$-$4$-$4$}	&$70$	&$560$		&		\\
{\tt johnson$16$-$2$-$4$}	&$120$	&$1{,}680$	&		\\
{\tt johnson$32$-$2$-$4$}	&$496$	&$14{,}880$	&		\\
{\tt keller$4$}			&$171$	&$5{,}100$	&		\\
{\tt hamming$6$-$2$}		&$64$	&$192$		&		\\
{\tt hamming$6$-$4$}		&$64$	&$1{,}312$	&		\\
{\tt hamming$8$-$2$}		&$256$	&$1{,}024$	&		\\
{\tt hamming$8$-$4$}		&$256$	&$11{,}776$	&		\\
{\tt san$200$\_$0.7$\_1}	&$200$	&$5{,}970$	&$30$		\\
{\tt san$200$\_$0.7$\_2}	&$200$	&$5{,}970$	&$18$		\\
{\tt san$200$\_$0.9$\_1}	&$200$	&$1{,}990$	&$70$		\\
{\tt san$200$\_$0.9$\_2}	&$200$	&$1{,}990$	&$60$		\\
{\tt san$200$\_$0.9$\_3}	&$200$	&$1{,}990$	&$44$		\\
{\tt san$400$\_$0.5$\_1}	&$400$	&$39{,}900$	&$13$		\\
{\tt san$400$\_$0.7$\_1}	&$400$	&$23{,}940$	&$40$		\\
{\tt san$400$\_$0.7$\_2}	&$400$	&$23{,}940$	&$30$		\\
{\tt san$400$\_$0.7$\_3}	&$400$	&$23{,}940$	&$22$		\\
{\tt san$400$\_$0.9$\_1}	&$400$	&$7{,}980$	&$100$		\\
\hline
\end{tabular}

\label{graphs1}
\end{table}

\begin{table}[t]
\centering
\caption{Benchmark graphs (continued from Table \ref{graphs1}).}
\begin{tabular}{lrrr}
$\bar G$			&$n$	&$m$		&$\alpha(G)$	\\
\hline
{\tt sanr$200$\_$0.7$}		&$200$	&$6{,}032$	&		\\
{\tt sanr$200$\_$0.9$}		&$200$	&$2{,}037$	&		\\
{\tt sanr$400$\_$0.5$}		&$400$	&$39{,}816$	&		\\
{\tt sanr$400$\_$0.7$}		&$400$	&$23{,}931$	&		\\
{\tt brock$200$\_1}		&$200$	&$5{,}066$	&$21$		\\
{\tt brock$200$\_2}		&$200$	&$10{,}024$	&$12$		\\
{\tt brock$200$\_3}		&$200$	&$7{,}852$	&$15$		\\
{\tt brock$200$\_4}		&$200$	&$6{,}811$	&$17$		\\
{\tt brock$400$\_1}		&$400$	&$20{,}077$	&$27$		\\
{\tt brock$400$\_2}		&$400$	&$20{,}014$	&$29$		\\
{\tt brock$400$\_3}		&$400$	&$20{,}119$	&$31$		\\
{\tt brock$400$\_4}		&$400$	&$20{,}035$	&$33$		\\
{\tt p\_hat$300$-1}		&$300$	&$33{,}917$	&		\\
{\tt p\_hat$300$-2}		&$300$	&$22{,}922$	&		\\
{\tt p\_hat$300$-3}		&$300$	&$11{,}460$	&		\\
{\tt p\_hat$500$-1}		&$500$	&$93{,}181$	&		\\
{\tt p\_hat$500$-2}		&$500$	&$61{,}804$	&		\\
{\tt p\_hat$500$-3}		&$500$	&$30{,}950$	&		\\
{\tt MANN\_a9}			&$45$	&$72$		&		\\
{\tt MANN\_a27}			&$378$	&$702$		&		\\
\hline
\end{tabular}

\label{graphs2}
\end{table}

\section{Experimental results}\label{results}

Henceforth, we refer as {\sc WAO} to the algorithm that results from the
formulation of Section~\ref{formul}. This denomination is an acronym after
``Widest Acyclic Orientation,'' as by (\ref{alphafromomega}) what is sought
during the evolutionary search is an acyclic orientation whose minimum chain
decomposition has the most chains over $\Omega(G)$.

{\sc WAO} iterates for $g$ generations, each one characterized by a population
of fixed size $s$. After generating the last population, it outputs the best
individual found during the entire evolutionary search. For $k>1$, the $k$th
population is obtained from the $k-1$st population as follows. First an elitist
step is performed, resulting in the transfer of the $fs$ fittest individuals
from the current population to the new, with $0\le f<1$. Then {\sc WAO} performs
the following iteration until the new population is full: with probability
$p_c$, two individuals are selected from the current population and the
crossover operator is applied to them, the resulting two individuals being then
added to the new population; with probability $1-p_c$, one single individual is
selected from the current population and then is subjected to the mutation
operator before being added to the new population.

In order to decide on an appropriate selection method, we ran several initial
experiments on reasonably-sized graphs. From these experiments emerged not only
the selection method of our choice but also the suite of parameters we would
adopt in all further experiments (we discuss these later). The selection method
we used in our experiments picks individuals proportionally to their linearly
normalized fitness in the current population. For $1\le k\le s$, this means that
the $k$th fittest individual---that is, $\omega$ such that $f(\omega)$ is the
$k$th greatest---is selected with probability proportional to
\begin{equation}
g(\omega)=L-\left(\frac{L-1}{s-1}\right)(k-1).
\label{normfitness}
\end{equation}
Ties between two individuals are broken by taking the individual that was added
to the current population first as the fitter one. In (\ref{normfitness}), $L$
is the factor by which the linearly normalized fitness of the fittest individual
in the current population (the $k=1$ case) is greater than that of the least
fit individual (the $k=s$ case); that is, $L=g(\omega_1)/g(\omega_s)$, where
$\omega_1$ and $\omega_s$ are those two individuals, respectively.

We present our results in comparison to those obtained by the heuristics of the
DIMACS challenge \cite{jt96}. This is not to say that the best results known to
date are necessarily the ones obtained by those heuristics, since several new
methods have appeared in the meantime for the maximum independent set problem
under one of its guises (e.g., \cite{bbp02,bbpr02} and their references).
However, all those more recent methods invariably go back to the DIMACS
challenge heuristics as references for comparison, so those heuristics serve as
an indirect basis for other comparisons as well. One exception to this
comparison rule we have adopted is the genetic-algorithm approach of
\cite{aot97}, which, like {\sc WAO}, employs a nontrivial crossover operator. We
present next a brief description of all the eleven heuristics to which we
compare {\sc WAO} directly.

\bigskip
\noindent
{\sc B\&C} \cite{bccp96}. This is a branch-and-cut method for which cutting
planes are generated based on the more general technique of \cite{bcc93}.
It starts with an integer programming formulation of the maximum clique
problem, and proceeds from the initial relaxation by generating new cutting
planes and incorporating them into the current linear program.

\bigskip
\noindent
{\sc Cliqmerge} \cite{bn96}. This heuristic is based on a procedure that finds
a maximum clique in the subgraph induced by the nodes of two cliques when they
are merged together. The essence of the method is to find a maximum bipartite
matching in the complement of this subgraph.

\bigskip
\noindent
{\sc Squeeze} \cite{bglm96}. This is a branch-and-bound algorithm for the
maximum independent set problem. Its lower bounds are obtained through a
reduction to the problem of minimizing a general quadratic $0$-$1$ function.

\bigskip
\noindent
{\sc CBH} \cite{ghp96}. This is an interior-point approach (cf.\ \cite{g92})
to the determination of maximum independent sets. Following a continuous
formulation of the problem, a relaxation of it is solved and the resulting
solution is rounded by a heuristic based on \cite{cp90,b91}.

\bigskip
\noindent
{\sc RB-clique} \cite{gr96}. This method uses backtracking ``coordinates''
as the entities on which restricted backtracking is to be applied while seeking
a maximum clique. The restrictions to which the backtracking coordinates are
subject are given as input.

\bigskip
\noindent
{\sc AtA} \cite{g96}. This strategy employs recurrent neural networks
(cf.\ \cite{b93}) to find maximum cliques. The crux of the approach is an
adaptive procedure for the determination of appropriate threshold parameters
and initial state for the neural network.

\bigskip
\noindent
{\sc SA\&GH} \cite{hp96}. This is a blend of heuristics to find maximum cliques.
In most cases it employs simply simulated annealing \cite{kgv83}, but for very
dense graphs the greedy heuristic of \cite{j74} is used.

\bigskip
\noindent
{\sc XSD} \cite{jsg96}. This is a family of heuristic methods to find maximum
cliques. The methods are all related to neural-network models and include
deterministic and stochastic descent approaches, with or without an intervening
learning step between restarts.

\bigskip
\noindent
{\sc B\&B} \cite{ms96}. This is a branch-and-bound approach to the maximum
independent set problem. Upper bounds are obtained through a procedure derived
from edge projection, a specialization of the clique projection of \cite{lp86}.

\bigskip
\noindent
{\sc XT} \cite{sg96}. This is a family of three variants of tabu search
\cite{gl97}, two deterministic and one probabilistic.

\bigskip
\noindent
{\sc OCH} \cite{aot97}. This is a genetic algorithm for the maximum independent
set problem. Each individual is an $n$-digit binary number, each digit
indicating whether the corresponding node is in the independent set or not. The
centerpiece of the method is the so-called optimized crossover, which generates
one optimal offspring based on the same merging procedure of {\sc Cliqmerge} and
one other having a random character.

\bigskip
All the experiments we report on were conducted with $g=10n$, $s=1.5n$,
$f=0.05$, $p_c=0.2$, and $L\in\{15,30\}$. As we indicated earlier, these reflect
policies and values that emerged from early experiments on reasonably-sized
graphs. They by no means represent optimal decisions of any sort, since the
number of possible choices is, naturally, far too large.

In all experiments, we also made use of the maximum-flow code of
\cite{hipr.tar-url}, which implements the algorithm of \cite{cg97}, to compute
the value of $f(\omega)$ as explained in Section \ref{formul}. As we also
indicated in that section, it suffices for the maximum-flow computation to stop
right after completing its first phase, since the value of the maximum flow, and
hence the size of the independent set that the individual contributes, is
already known at this point. We also recall that running the maximum-flow code
through its first phase only is sufficient even for the best individual found
during the whole evolutionary search. In this case, what we need is to enumerate
the members of the independent set contributed by that individual, not simply
to know its size, but this can be obtained by examining the minimum cut that is
also already known as the first phase ends.

\begin{figure}
\centering
\begin{tabular}{c@{\hspace{0.10in}}c}
\scalebox{0.30}{\includegraphics{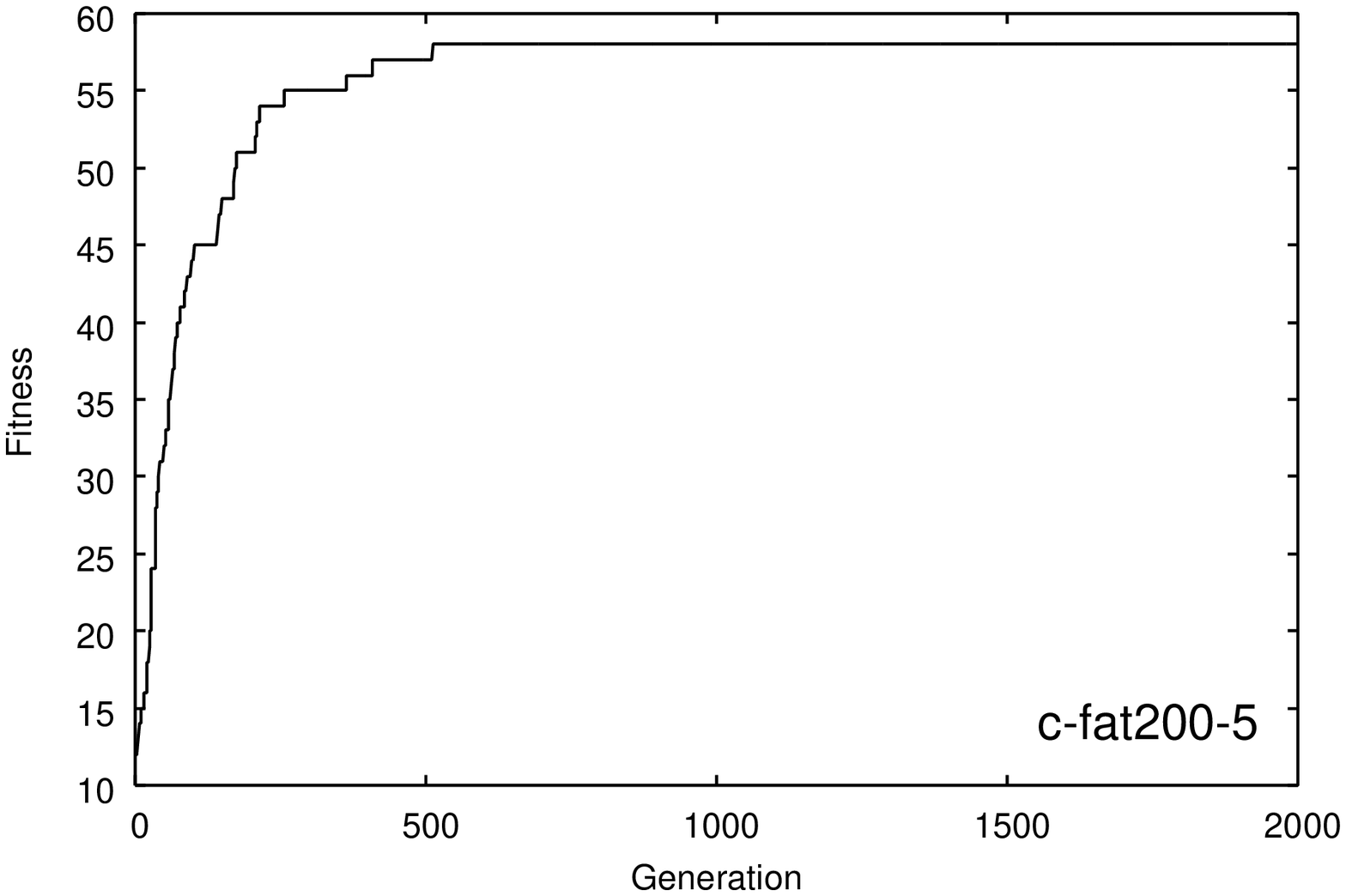}}&
\scalebox{0.30}{\includegraphics{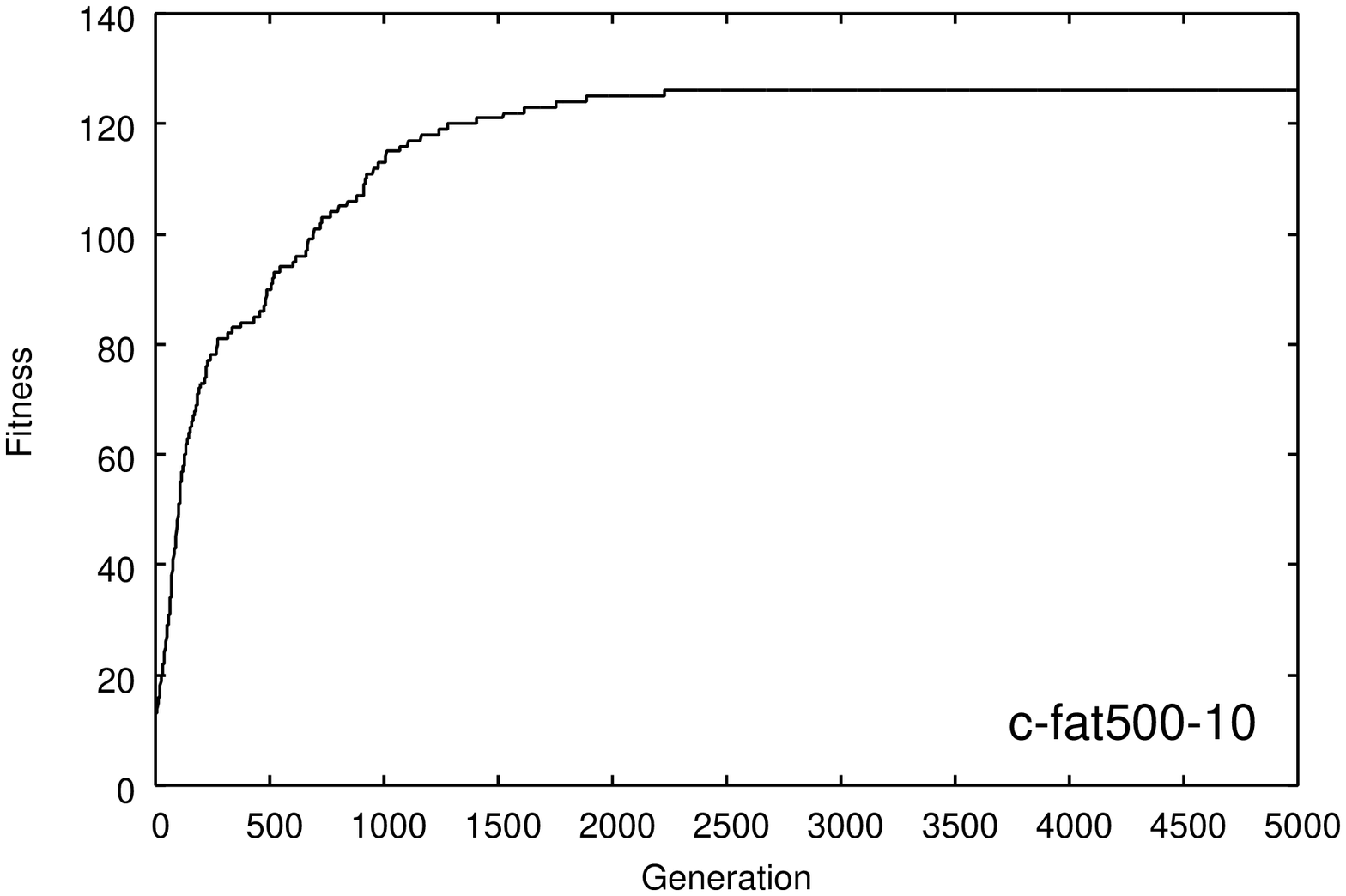}}\\
\scalebox{0.30}{\includegraphics{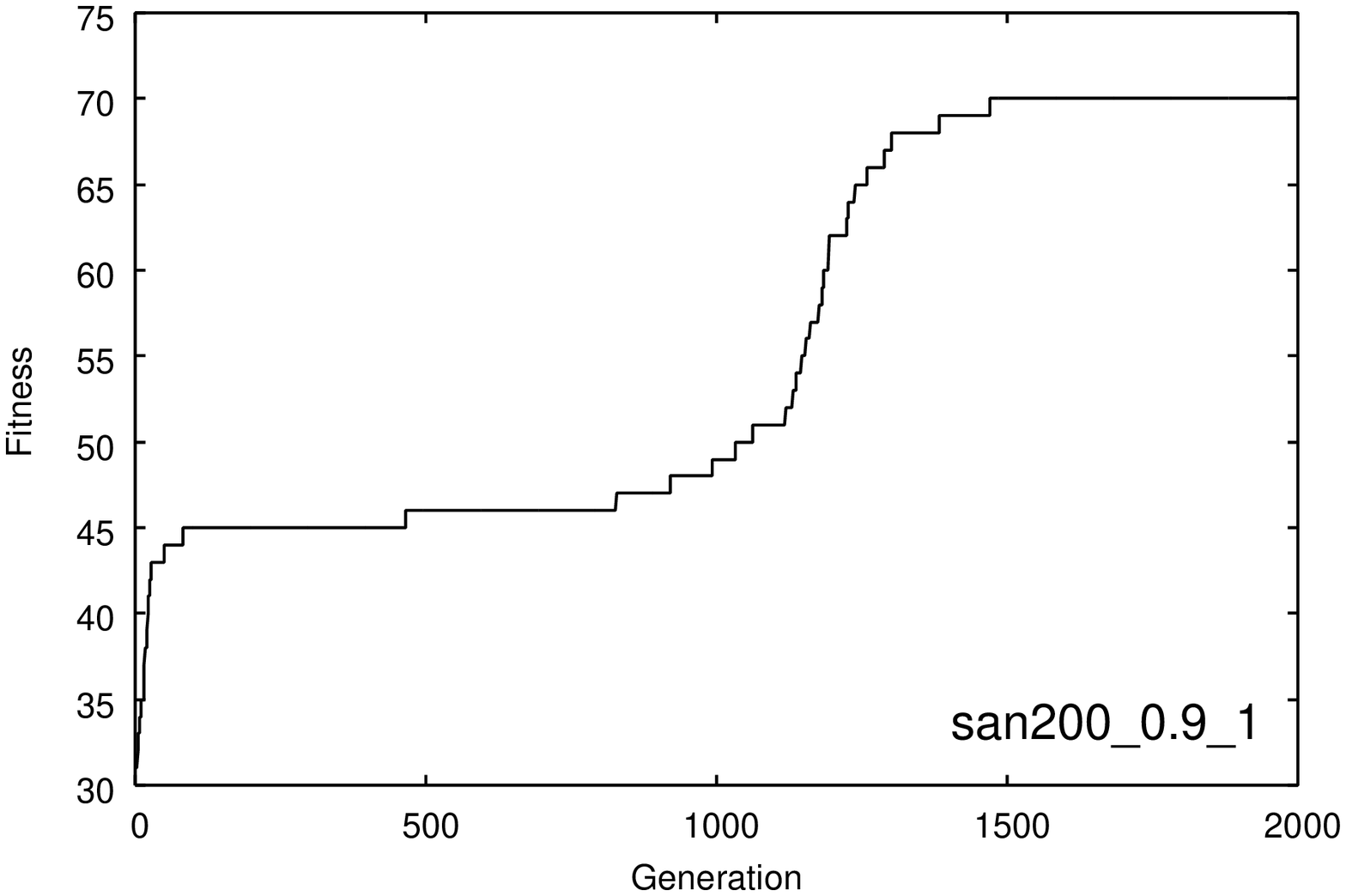}}&
\scalebox{0.30}{\includegraphics{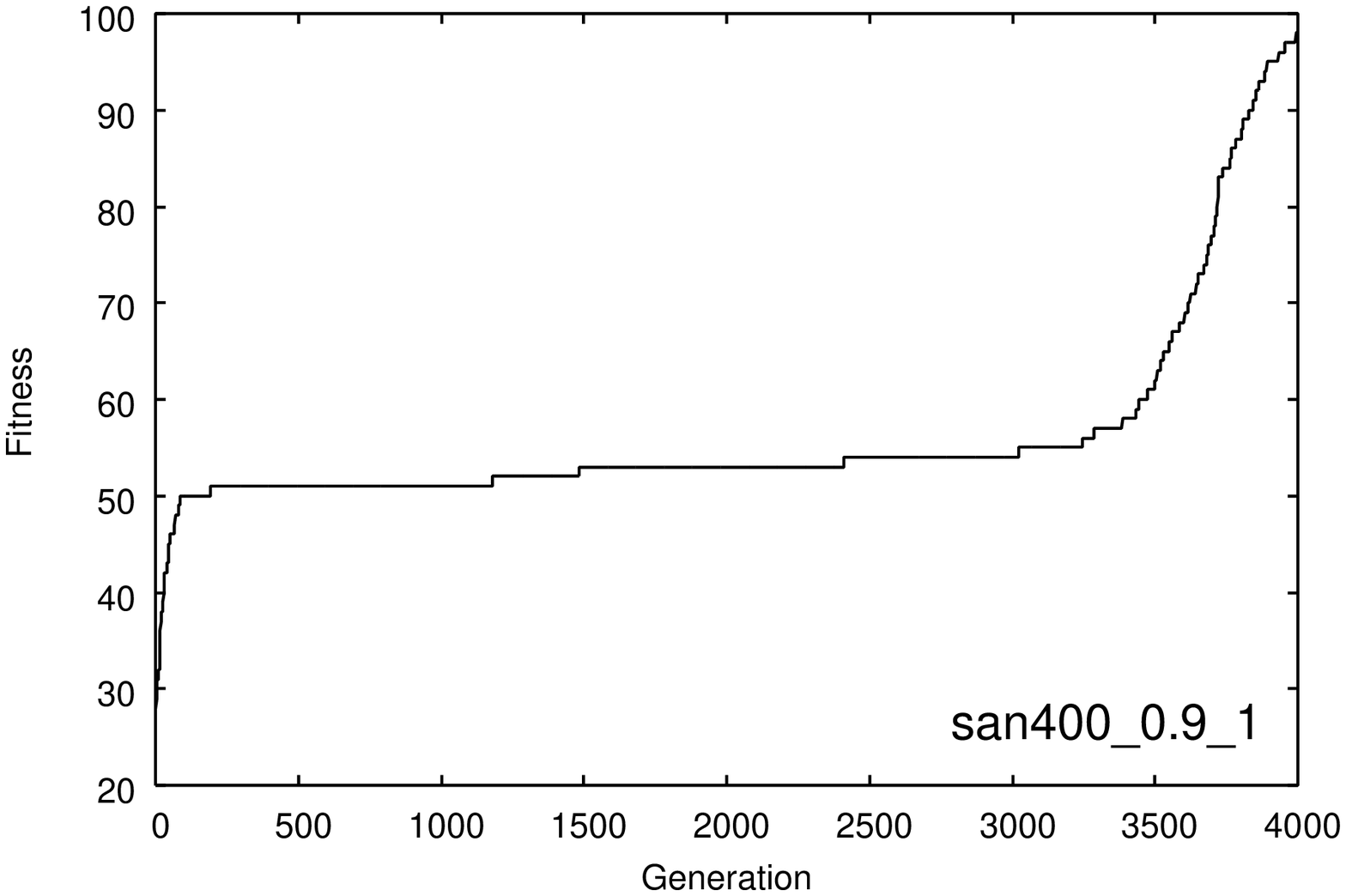}}\\
\scalebox{0.30}{\includegraphics{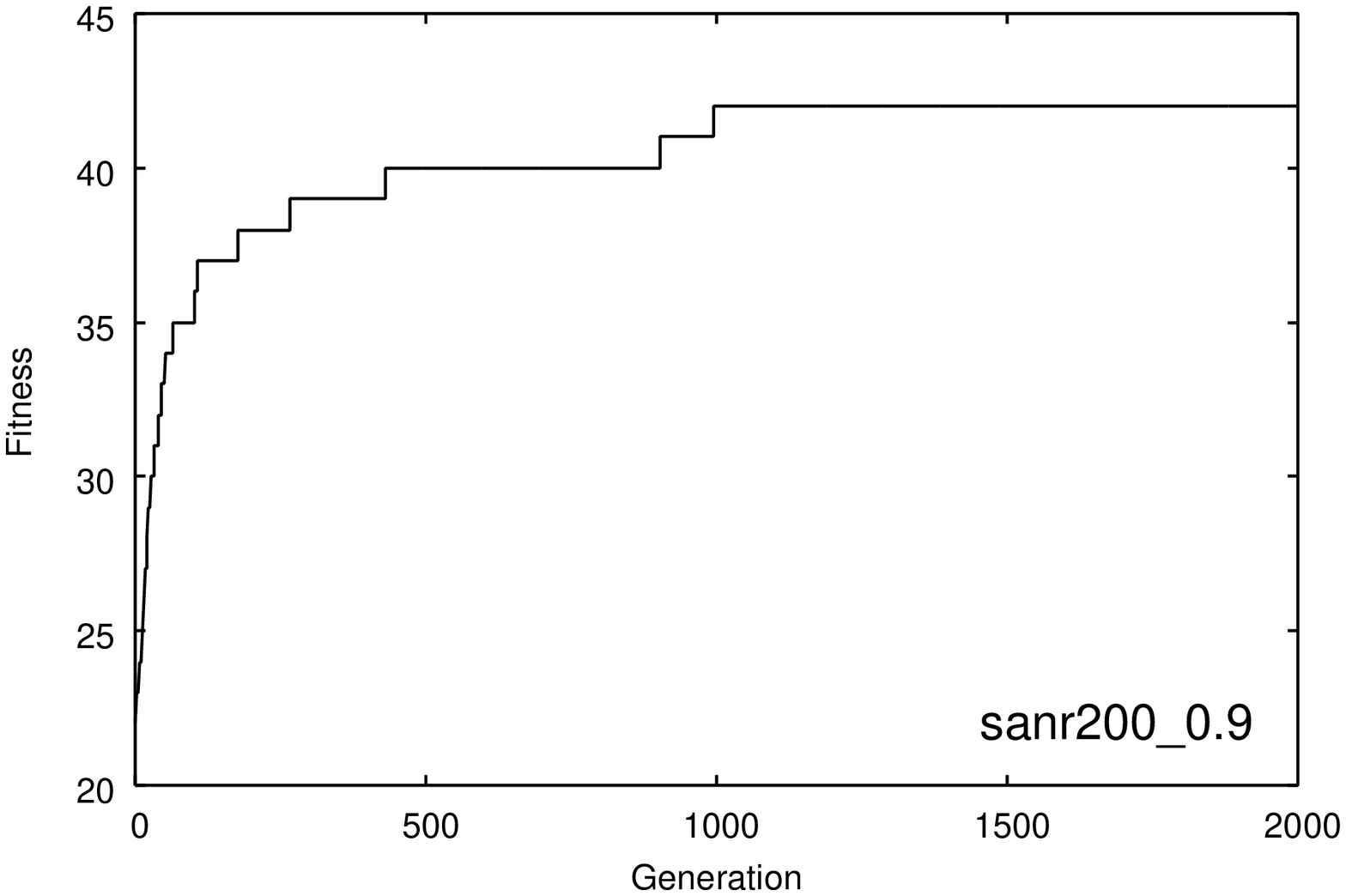}}&
\scalebox{0.30}{\includegraphics{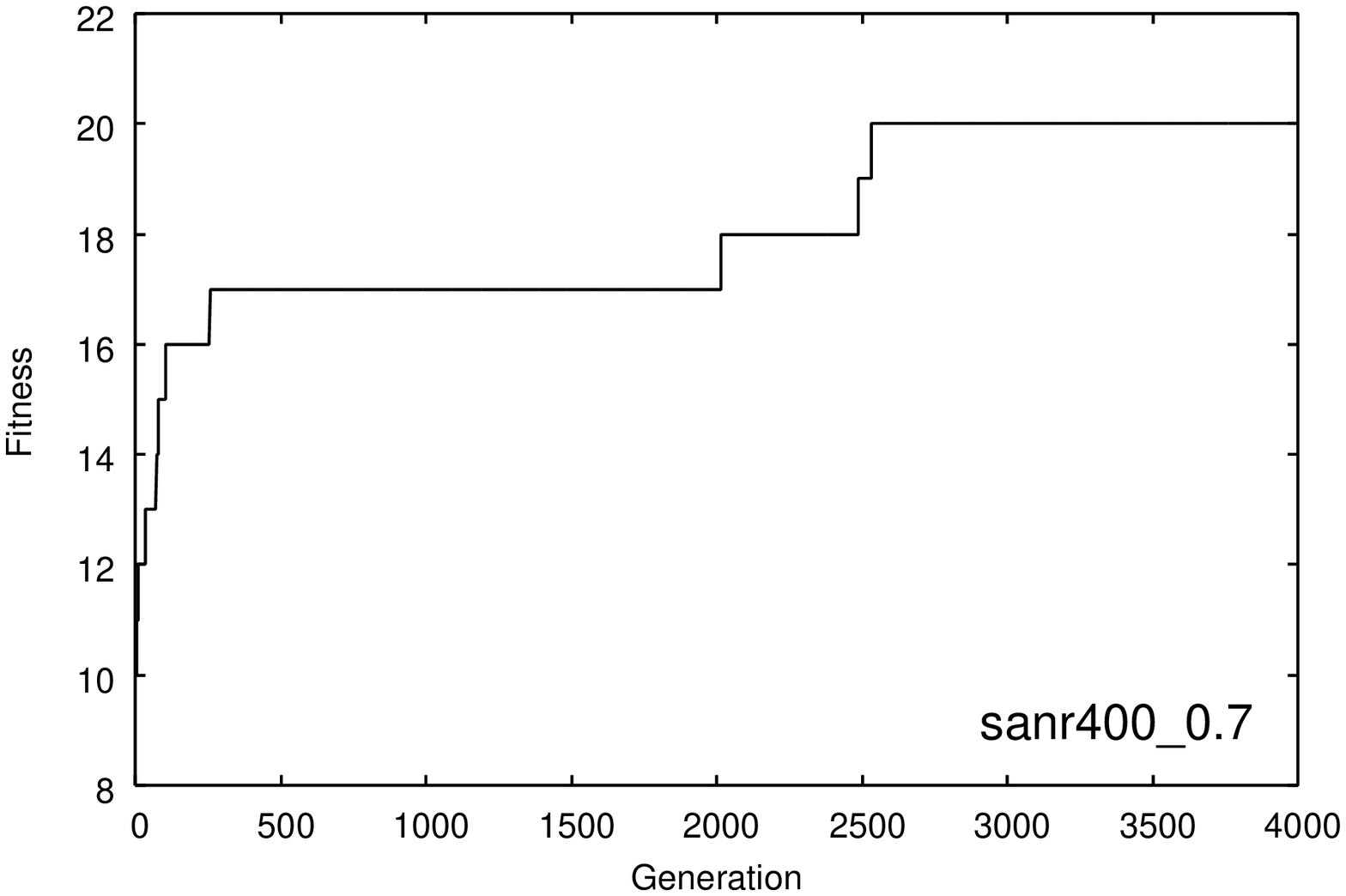}}\\
\scalebox{0.30}{\includegraphics{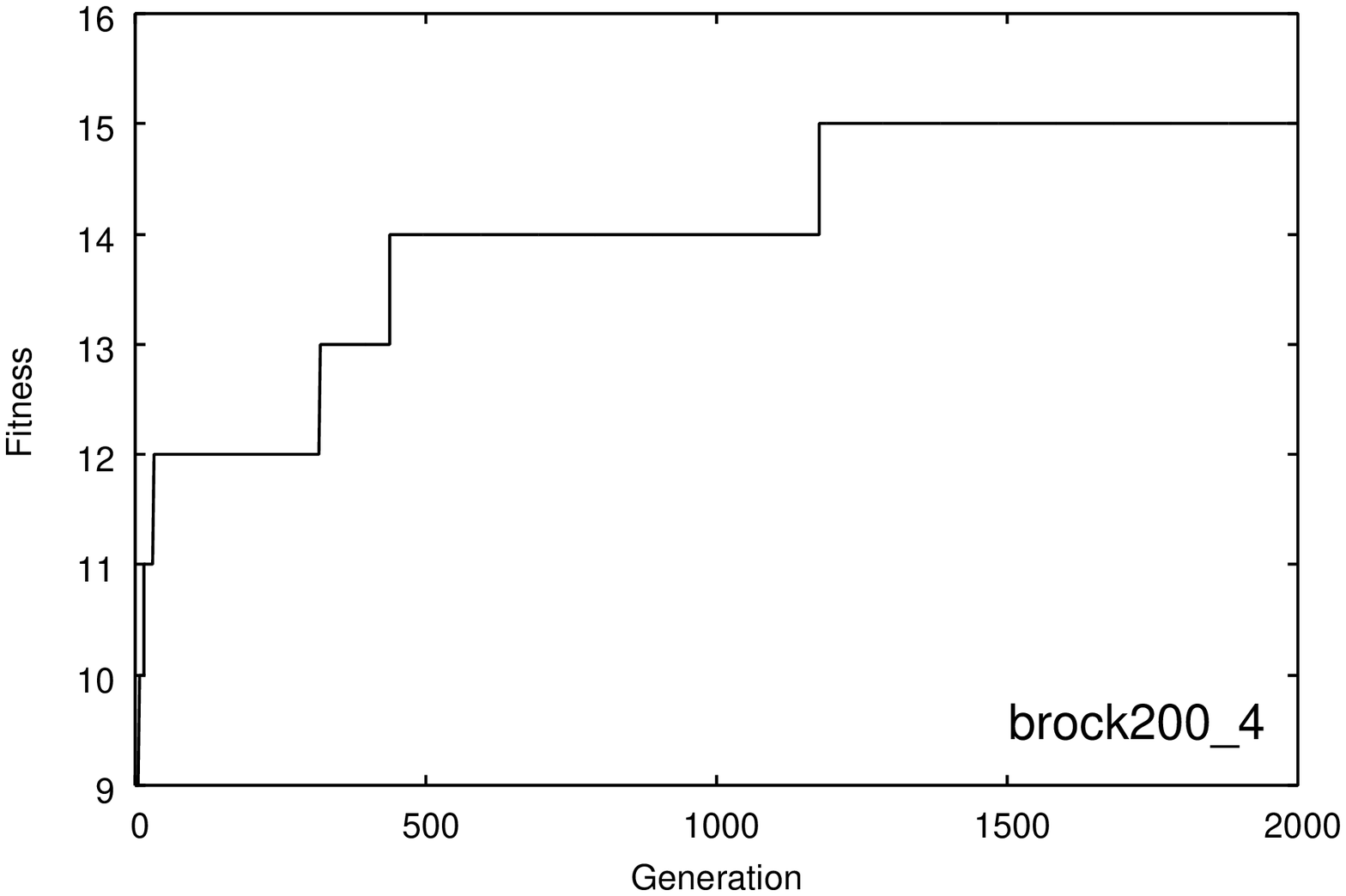}}&
\scalebox{0.30}{\includegraphics{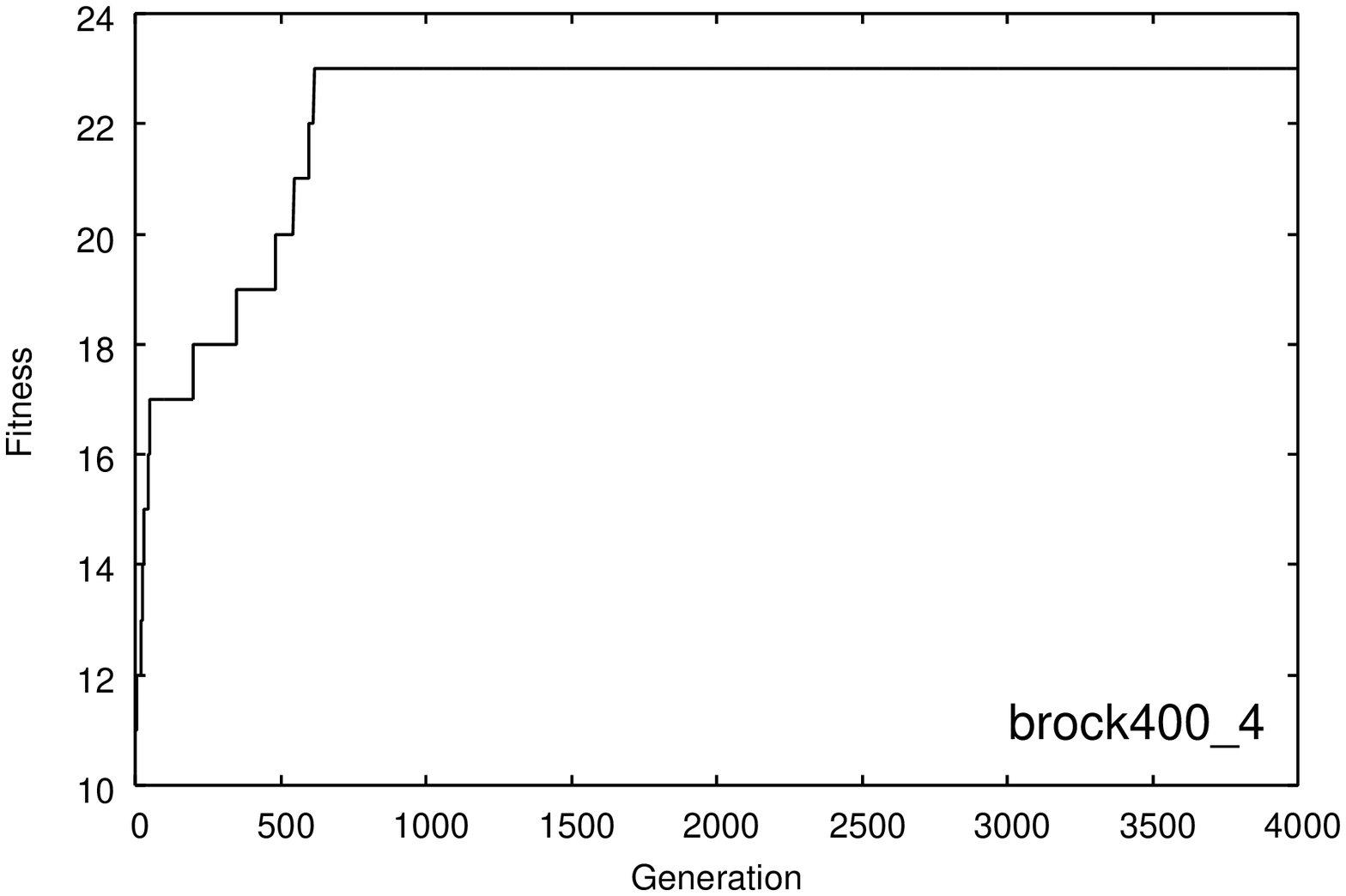}}\\
\end{tabular}
\caption{Fitness evolution for selected graphs.}
\label{fitnessplots1}
\end{figure}

\begin{figure}
\centering
\begin{tabular}{c@{\hspace{0.10in}}c}
\scalebox{0.30}{\includegraphics{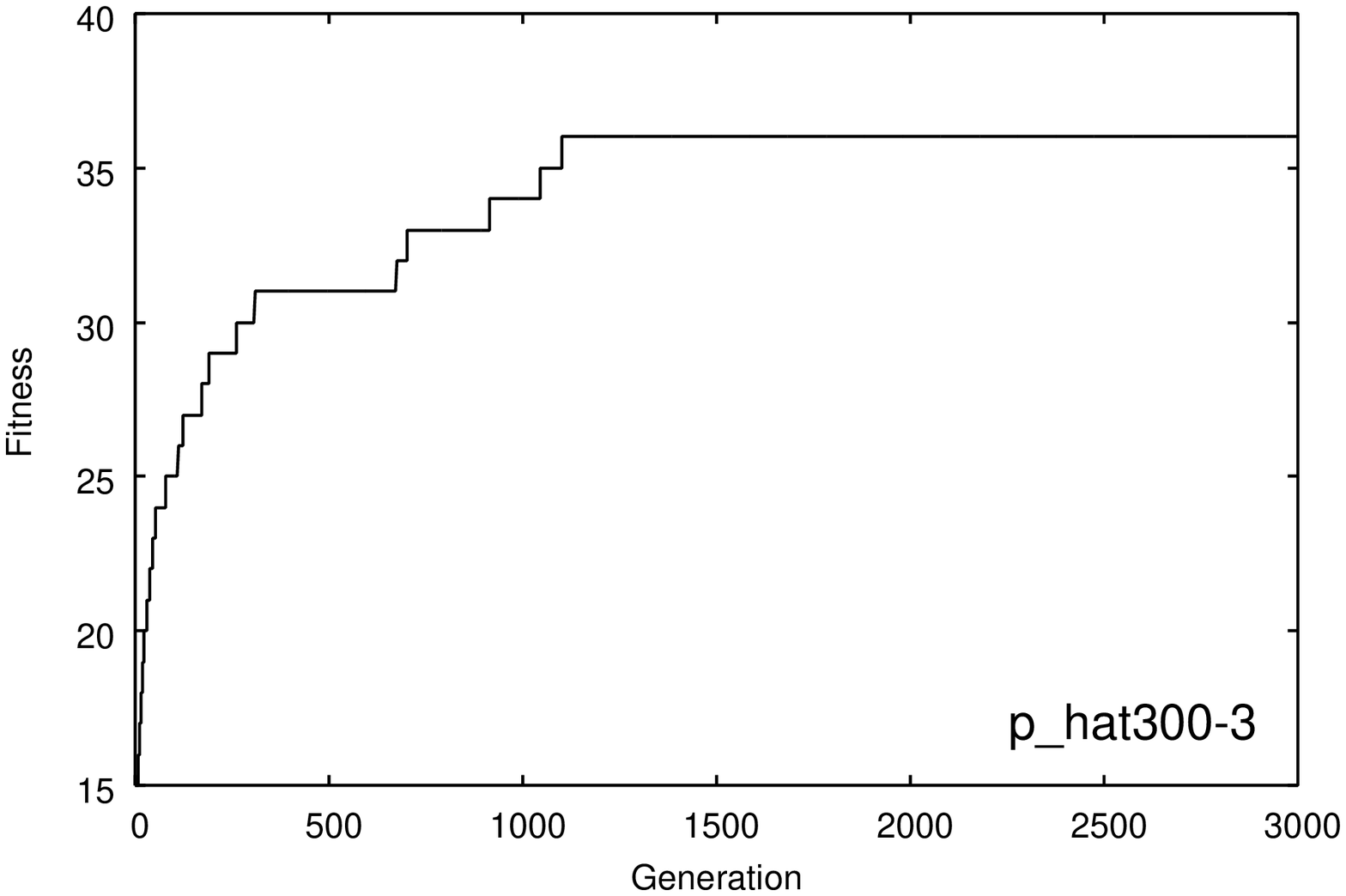}}&
\scalebox{0.30}{\includegraphics{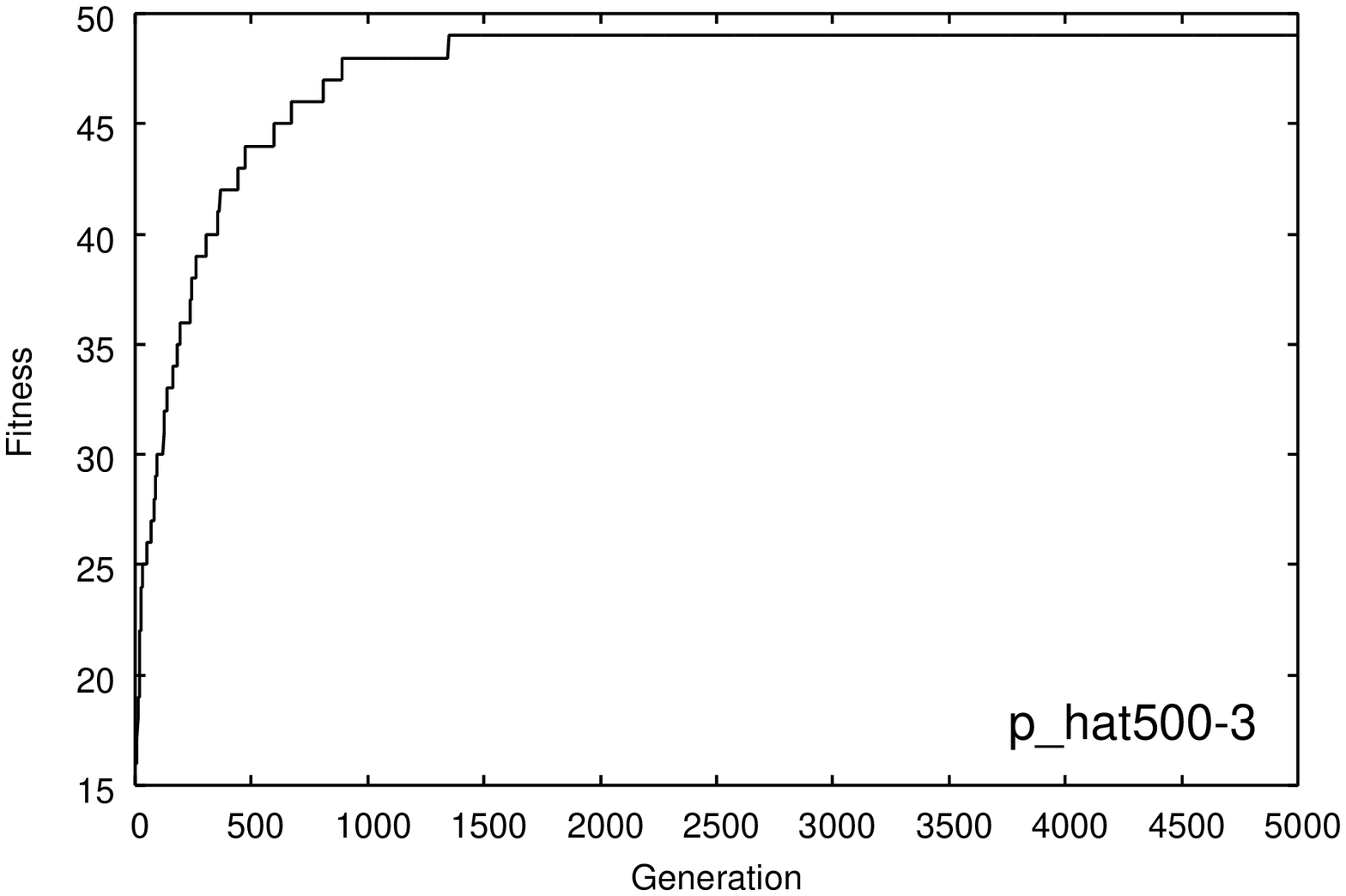}}\\
\end{tabular}
\caption{Fitness evolution for selected graphs (continued from Figure
\ref{fitnessplots1}).}
\label{fitnessplots2}
\end{figure}

Our results are shown in Figures \ref{fitnessplots1} and \ref{fitnessplots2},
and in Tables \ref{results1a} through \ref{results2c}. The two figures show, for
ten graphs selected from Tables \ref{graphs1} and \ref{graphs2}, the evolution
of the best fitness (as given by (\ref{fitnessfromomega})) ever found for an
individual as the generations elapse during the best of twenty independent runs
(the one that eventually yielded the largest independent set for that graph). In
other words, they show the size of the largest independent set yet identified.
These ten graphs were selected because, in terms of what is shown in Figures
\ref{fitnessplots1} and \ref{fitnessplots2}, they have led the evolutionary
search to behave either in a way that we found to be somewhat typical or a way
that yields interesting insight. We return to this shortly.

Tables \ref{results1a} through \ref{results2c} show the results obtained by
{\sc WAO} alongside the results of the competing algorithms we outlined earlier
in this section. Tables \ref{results1a}--\ref{results1c} refer to the graphs in
Table \ref{graphs1}, while Tables \ref{results2a}--\ref{results2c} refer to
those in Table \ref{graphs2}. For each graph $\bar G$, the tables give the value
of $\alpha(G)$, when known from design characteristics, and the sizes of the
maximum independent sets obtained by the algorithms on $G$, when available. The
result reported for {\sc WAO} on each graph is the best result found over twenty
independent runs. For the other methods, the results reported are the best
results they yielded, as published in \cite{jt96}. The number appearing in
parentheses next to ``{\sc WAO}'' in Tables \ref{results1c} and \ref{results2c}
is the value of $L$ that was used to obtain the results listed in the
corresponding column.

The several runs of {\sc WAO} were executed on a relatively wide assortment of
machine architectures, so we refrain from providing detailed timing data. Also,
comparing running times to those of the other methods---obtained roughly one
decade ago---would be cumbersome however we tampered with the numbers seeking to
compensate for the technological gap. Given these constraints, all we do is
mention that each of our runs tended to complete somewhere between very few
seconds and a week, depending on the graph at hand.

{\sc WAO} is a competitive method, by all that can be inferred from Tables
\ref{results1a}--\ref{results2c}. We have in the tables used a bold typeface to
indicate the instances on which {\sc WAO} performed at least as well as the best
performers. Several entries are thus marked, and for many that are not {\sc WAO}
is seen to have fallen short by a very narrow margin.

Returning to the plots in Figures \ref{fitnessplots1} and \ref{fitnessplots2}
may highlight some of the patterns that help explain success or failure at
meeting the best performers' figures during our experiments with {\sc WAO}.
Of the ten graphs to which those figures refer, three correspond to cases in
which {\sc WAO} missed by a narrow margin ({\tt san$400$\_$0.9$\_1},
{\tt sanr$400$\_$0.7$}, and {\tt brock$200$\_4}) and one to a case in which it
missed widely ({\tt brock$400$\_4}). What seems to distinguish one group from
the other is that in the latter case the plot becomes flat early in the
evolution, perhaps signaling an inherent hardness at escaping some
particularly difficult local maximum. For the three graphs in the former group,
however, and in fact for the six graphs in the figures outside either group,
evolution seems to lead to fitness growth more or less steadily along a
comparatively larger number of generations, even though for the {\tt san} graphs
it nearly stalls for a significant number of generations before it gains
momentum again. In the case of {\tt san$400$\_$0.9$\_1}, particularly, it
appears quite likely that a few more generations would have bridged the narrow
gap between the $98$ that {\sc WAO} achieved and the $100$ of its best
contenders.

\section{Concluding remarks}\label{concl}

We have in this paper introduced {\sc WAO}, a novel evolutionary heuristic for
the maximum independent set problem. {\sc WAO} is based on a view of the problem
that relates independent sets to the acyclic orientations of the graph, and
seeks to identify an acyclic orientation that is widest (has the decomposition
into the fewest number of chains that requires the most chains) over the set of
all the acyclic orientations of the graph. It incorporates no additional
sophistication into the usual evolutionary-algorithm machinery, but rather into
the design of the individuals' representations and the evolutionary operators,
all based on complex graph-theoretic notions.

We have found our new heuristic to perform competitively when compared to
several others on the DIMACS benchmark graphs. Notwithstanding this, there
certainly is room for further investigation and improvements. For example, there
may exist a better set of parameters for the evolutionary search, just as there
may exist a better functional dependence of $g$ and $s$ on $n$ (and perhaps also
on $m$, unlike what we adopted in our experiments). Likewise, it is also
conceivable that the formulation itself may be improved by the incorporation of
optimizations into the crossover or mutation operator, or even by the
introduction of new operators.

\subsection*{Acknowledgments}

The authors acknowledge partial support from CNPq, CAPES, the PRONEX initiative
of Brazil's MCT under contract 41.96.0857.00, and a FAPERJ BBP grant.

\clearpage

\clearpage

\begin{table}
\centering
\caption{Comparative performance on the benchmark graphs of
Table \ref{graphs1}.}
\begin{tabular}{lrrrrr}
$\bar G$			&$\alpha(G)$ &{\sc B\&C} &{\sc Cliqmerge} &{\sc Squeeze} &{\sc CBH} \\
\hline
{\tt c-fat$200$-$1$}		& &$12$ & &$12$ &$12$ \\
{\tt c-fat$200$-$2$}		& &$24$ & &$24$ &$24$ \\
{\tt c-fat$200$-$5$}		& &$58$ & &$58$ &$58$ \\
{\tt c-fat$500$-$1$}		& & & &$14$ &$14$ \\
{\tt c-fat$500$-$2$}		& & & &$26$ &$26$ \\
{\tt c-fat$500$-$5$}		& & & &$64$ &$64$ \\
{\tt c-fat$500$-$10$}		& & & &$126$ &$126$ \\
{\tt johnson$8$-$2$-$4$}	& &$4$ & &$4$ &$4$ \\
{\tt johnson$8$-$4$-$4$}	& &$14$ & &$14$ &$14$ \\
{\tt johnson$16$-$2$-$4$}	& &$8$ & &$8$ &$8$ \\
{\tt johnson$32$-$2$-$4$}	& & & &$16$ &$16$ \\
{\tt keller$4$}			& &$11$ &$11$ &$11$ &$10$ \\
{\tt hamming$6$-$2$}		& &$32$ & &$32$ &$32$ \\
{\tt hamming$6$-$4$}		& &$4$ & &$4$ &$4$ \\
{\tt hamming$8$-$2$}		& &$128$ & &$128$ &$128$ \\
{\tt hamming$8$-$4$}		& &$16$ &$16$ &$16$ &$16$ \\
{\tt san$200$\_$0.7$\_1}	&$30$ &$30$ & &$30$ &$15$ \\
{\tt san$200$\_$0.7$\_2}	&$18$ &$18$ & &$18$ &$12$ \\
{\tt san$200$\_$0.9$\_1}	&$70$ &$70$ & &$70$ &$46$ \\
{\tt san$200$\_$0.9$\_2}	&$60$ &$60$ & &$60$ &$36$ \\
{\tt san$200$\_$0.9$\_3}	&$44$ &$44$ & &$44$ &$30$ \\
{\tt san$400$\_$0.5$\_1}	&$13$ &$13$ & & &$8$ \\
{\tt san$400$\_$0.7$\_1}	&$40$ & & & &$20$ \\
{\tt san$400$\_$0.7$\_2}	&$30$ &$30$ & & &$15$ \\
{\tt san$400$\_$0.7$\_3}	&$22$ & & & &$14$ \\
{\tt san$400$\_$0.9$\_1}	&$100$ & & &$100$ &$50$ \\
\hline
\end{tabular}

\label{results1a}
\end{table}

\begin{table}
\centering
\caption{Comparative performance on the benchmark graphs of
Table \ref{graphs1} (continued from Table \ref{results1a}).}
\begin{tabular}{lrrrrr}
$\bar G$			&$\alpha(G)$ &{\sc RB-clique} &{\sc AtA} &{\sc SA\&GH} &{\sc XSD} \\
\hline
{\tt c-fat$200$-$1$}		& & & & &$12$ \\
{\tt c-fat$200$-$2$}		& & & & &$24$ \\
{\tt c-fat$200$-$5$}		& & & & &$58$ \\
{\tt c-fat$500$-$1$}		& & & & &$14$ \\
{\tt c-fat$500$-$2$}		& & & & &$26$ \\
{\tt c-fat$500$-$5$}		& & & & &$64$ \\
{\tt c-fat$500$-$10$}		& & & & &$126$ \\
{\tt johnson$8$-$2$-$4$}	& & & & &$4$ \\
{\tt johnson$8$-$4$-$4$}	& & & & &$14$ \\
{\tt johnson$16$-$2$-$4$}	& & & & &$8$ \\
{\tt johnson$32$-$2$-$4$}	& & & & &$16$ \\
{\tt keller$4$}			& &$11$ &$11$ &$11$ &$11$ \\
{\tt hamming$6$-$2$}		& & & & &$32$ \\
{\tt hamming$6$-$4$}		& & & & &$4$ \\
{\tt hamming$8$-$2$}		& & & & &$128$ \\
{\tt hamming$8$-$4$}		& &$16$ &$16$ &$16$ &$16$ \\
{\tt san$200$\_$0.7$\_1}	&$30$ & & & &$30$ \\
{\tt san$200$\_$0.7$\_2}	&$18$ & & & &$15$ \\
{\tt san$200$\_$0.9$\_1}	&$70$ & & & &$70$ \\
{\tt san$200$\_$0.9$\_2}	&$60$ & & & &$60$ \\
{\tt san$200$\_$0.9$\_3}	&$44$ & & & &$36$ \\
{\tt san$400$\_$0.5$\_1}	&$13$ & & & &$9$ \\
{\tt san$400$\_$0.7$\_1}	&$40$ & & & &$33$ \\
{\tt san$400$\_$0.7$\_2}	&$30$ & & & &$19$ \\
{\tt san$400$\_$0.7$\_3}	&$22$ & & & &$16$ \\
{\tt san$400$\_$0.9$\_1}	&$100$ & & & &$100$ \\
\hline
\end{tabular}

\label{results1b}
\end{table}

\begin{table}
\centering
\caption{Comparative performance on the benchmark graphs of
Table \ref{graphs1} (continued from Table \ref{results1b}).}
\begin{tabular}{lrrrrrr}
$\bar G$			&$\alpha(G)$ &{\sc B\&B} &{\sc XT} &{\sc OCH} &{\sc WAO} ($15$)&{\sc WAO} ($30$) \\
\hline
{\tt c-fat$200$-$1$}		& & & &$12$ &$\bf 12$ &$\bf 12$ \\
{\tt c-fat$200$-$2$}		& & & &$24$ &$\bf 24$ &$\bf 24$ \\
{\tt c-fat$200$-$5$}		& & & &$58$ &$\bf 58$ &$\bf 58$ \\
{\tt c-fat$500$-$1$}		& & & &$14$ &$\bf 14$ &$\bf 14$ \\
{\tt c-fat$500$-$2$}		& & & &$26$ &$\bf 26$ &$\bf 26$ \\
{\tt c-fat$500$-$5$}		& & & &$64$ &$\bf 64$ &$\bf 64$ \\
{\tt c-fat$500$-$10$}		& & & &$126$ &$\bf 126$ &$\bf 126$ \\
{\tt johnson$8$-$2$-$4$}	& & & &$4$ &$\bf 4$ &$\bf 4$ \\
{\tt johnson$8$-$4$-$4$}	& & & &$14$ &$\bf 14$ &$\bf 14$ \\
{\tt johnson$16$-$2$-$4$}	& & & &$8$ &$\bf 8$ &$\bf 8$ \\
{\tt johnson$32$-$2$-$4$}	& & & &$16$ &$\bf 16$ &$\bf 16$ \\
{\tt keller$4$}			& &$11$ &$11$ &$11$ &$\bf 11$ &$\bf 11$ \\
{\tt hamming$6$-$2$}		& & & &$32$ &$\bf 32$ &$\bf 32$ \\
{\tt hamming$6$-$4$}		& & & &$4$ &$\bf 4$ &$\bf 4$ \\
{\tt hamming$8$-$2$}		& & & &$128$ &$\bf 128$ &$\bf 128$ \\
{\tt hamming$8$-$4$}		& &$16$ &$16$ &$16$ &$\bf 16$ &$\bf 16$ \\
{\tt san$200$\_$0.7$\_1}	&$30$ & & &$30$ &$16$ &$16$ \\
{\tt san$200$\_$0.7$\_2}	&$18$ & & &$15$ &$14$ &$14$ \\
{\tt san$200$\_$0.9$\_1}	&$70$ & & &$70$ &$\bf 70$ &$\bf 70$ \\
{\tt san$200$\_$0.9$\_2}	&$60$ & & &$60$ &$\bf 60$ &$58$ \\
{\tt san$200$\_$0.9$\_3}	&$44$ & & &$36$ &$37$ &$\bf 44$ \\
{\tt san$400$\_$0.5$\_1}	&$13$ & & &$13$ &$8$ &$8$ \\
{\tt san$400$\_$0.7$\_1}	&$40$ & & &$40$ &$20$ &$20$ \\
{\tt san$400$\_$0.7$\_2}	&$30$ & & &$30$ &$17$ &$17$ \\
{\tt san$400$\_$0.7$\_3}	&$22$ & & &$16$ &$16$ &$16$ \\
{\tt san$400$\_$0.9$\_1}	&$100$ & & &$100$ &$54$ &$98$ \\
\hline
\end{tabular}

\label{results1c}
\end{table}

\begin{table}
\centering
\caption{Comparative performance on the benchmark graphs of
Table \ref{graphs2}.}
\begin{tabular}{lrrrrr}
$\bar G$			&$\alpha(G)$ &{\sc B\&C} &{\sc Cliqmerge} &{\sc Squeeze} &{\sc CBH} \\
\hline
{\tt sanr$200$\_$0.7$}		& & & &$18$ &$18$ \\
{\tt sanr$200$\_$0.9$}		& & & &$41$ &$41$ \\
{\tt sanr$400$\_$0.5$}		& & & & &$12$ \\
{\tt sanr$400$\_$0.7$}		& & & &$20$ &$20$ \\
{\tt brock$200$\_1}		&$21$ & & &$21$ &$20$ \\
{\tt brock$200$\_2}		&$12$ &$12$ &$11$ &$12$ &$12$ \\
{\tt brock$200$\_3}		&$15$ & & &$15$ &$14$ \\
{\tt brock$200$\_4}		&$17$ & &$16$ &$17$ &$16$ \\
{\tt brock$400$\_1}		&$27$ & & & &$23$ \\
{\tt brock$400$\_2}		&$29$ & &$25$ & &$24$ \\
{\tt brock$400$\_3}		&$31$ & & & &$23$ \\
{\tt brock$400$\_4}		&$33$ & &$25$ & &$24$ \\
{\tt p\_hat$300$-1}		& &$8$ &$8$ &$8$ &$8$ \\
{\tt p\_hat$300$-2}		& & &$25$ &$25$ &$25$ \\
{\tt p\_hat$300$-3}		& & &$36$ &$36$ &$36$ \\
{\tt p\_hat$500$-1}		& & & &$9$ &$9$ \\
{\tt p\_hat$500$-2}		& & & &$36$ &$35$ \\
{\tt p\_hat$500$-3}		& & & & &$49$ \\
{\tt MANN\_a9}			& & & &$16$ &$16$ \\
{\tt MANN\_a27}			& & &$126$ &$126$ &$121$ \\
\hline
\end{tabular}

\label{results2a}
\end{table}

\begin{table}
\centering
\caption{Comparative performance on the benchmark graphs of
Table \ref{graphs2} (continued from Table \ref{results2a}).}
\begin{tabular}{lrrrrr}
$\bar G$			&$\alpha(G)$ &{\sc RB-clique} &{\sc AtA} &{\sc SA\&GH} &{\sc XSD} \\
\hline
{\tt sanr$200$\_$0.7$}		& & & & &$18$ \\
{\tt sanr$200$\_$0.9$}		& & & & &$41$ \\
{\tt sanr$400$\_$0.5$}		& & & & &$12$ \\
{\tt sanr$400$\_$0.7$}		& & & & &$21$ \\
{\tt brock$200$\_1}		&$21$ & & & &$20$ \\
{\tt brock$200$\_2}		&$12$ &$12$ &$11$ &$11$ &$10$ \\
{\tt brock$200$\_3}		&$15$ & & & &$15$ \\
{\tt brock$200$\_4}		&$17$ &$17$ &$16$ &$16$ &$16$ \\
{\tt brock$400$\_1}		&$27$ & & & &$24$ \\
{\tt brock$400$\_2}		&$29$ &$25$ &$25$ &$25$ &$24$ \\
{\tt brock$400$\_3}		&$31$ & & & &$24$ \\
{\tt brock$400$\_4}		&$33$ &$33$ &$25$ &$25$ &$24$ \\
{\tt p\_hat$300$-1}		& &$8$ &$8$ &$8$ &$8$ \\
{\tt p\_hat$300$-2}		& &$25$ &$25$ &$25$ &$25$ \\
{\tt p\_hat$300$-3}		& &$35$ &$36$ &$36$ &$34$ \\
{\tt p\_hat$500$-1}		& & & & & \\
{\tt p\_hat$500$-2}		& & & & & \\
{\tt p\_hat$500$-3}		& & & & & \\
{\tt MANN\_a9}			& & & & & \\
{\tt MANN\_a27}			& &$126$ &$125$ &$126$ &$126$ \\
\hline
\end{tabular}

\label{results2b}
\end{table}

\begin{table}
\centering
\caption{Comparative performance on the benchmark graphs of
Table \ref{graphs2} (continued from Table \ref{results2b}).}
\begin{tabular}{lrrrrrr}
$\bar G$			&$\alpha(G)$ &{\sc B\&B} &{\sc XT} &{\sc OCH} &{\sc WAO} ($15$) &{\sc WAO} ($30$) \\
\hline
{\tt sanr$200$\_$0.7$}		& & & &$18$ &$17$ &$\bf 18$ \\
{\tt sanr$200$\_$0.9$}		& & & &$42$ &$\bf 42$ &$41$ \\
{\tt sanr$400$\_$0.5$}		& & & &$12$ &$11$ &$11$ \\
{\tt sanr$400$\_$0.7$}		& & & &$20$ &$20$ &$19$ \\
{\tt brock$200$\_1}		&$21$ & & &$21$ &$19$ &$19$ \\
{\tt brock$200$\_2}		&$12$ &$12$ &$11$ &$11$ &$10$ &$9$ \\
{\tt brock$200$\_3}		&$15$ & & &$14$ &$13$ &$13$ \\
{\tt brock$200$\_4}		&$17$ &$17$ &$16$ &$16$ &$15$ &$15$ \\
{\tt brock$400$\_1}		&$27$ & & &$24$ &$21$ &$22$ \\
{\tt brock$400$\_2}		&$29$ & &$25$ &$24$ &$21$ &$22$ \\
{\tt brock$400$\_3}		&$31$ & & &$24$ &$22$ &$22$ \\
{\tt brock$400$\_4}		&$33$ &$33$ &$25$ &$24$ &$23$ &$22$ \\
{\tt p\_hat$300$-1}		& &$8$ &$8$ &$8$ &$\bf 8$ &$7$ \\
{\tt p\_hat$300$-2}		& &$25$ &$25$ &$25$ &$24$ &$\bf 25$ \\
{\tt p\_hat$300$-3}		& &$36$ &$36$ &$36$ &$34$ &$\bf 36$ \\
{\tt p\_hat$500$-1}		& & & &$9$ &$\bf 9$ &$\bf 9$ \\
{\tt p\_hat$500$-2}		& & & &$36$ &$34$ &$35$ \\
{\tt p\_hat$500$-3}		& & & &$49$ &$\bf 49$ &$48$ \\
{\tt MANN\_a9}			& & & &$16$ &$\bf 16$ &$\bf 16$ \\
{\tt MANN\_a27}			& &$126$ &$125$ &$126$ &$\bf 126$ &$\bf 126$ \\
\hline
\end{tabular}

\label{results2c}
\end{table}

\clearpage


\end{document}